\documentclass{article}
\usepackage{spconf,amsmath,epsfig}
\usepackage{subcaption, booktabs, amsmath, hyperref}

\usepackage{pifont}

\usepackage{array}
\usepackage{placeins}

\usepackage{lineno,hyperref}
\title{Multi-Modal Deep Learning for Multi-Temporal Urban Mapping with a Partly Missing Optical Modality}
\name{Sebastian Hafner\textsuperscript{1,}\thanks{The research is funded by the Swedish National Space Agency, the project 'EO-AI4Urban' within the ESA and Chinese Ministry of Science and Technology's Dragon 5 Program, Digital Futures within the project 'EO-AI4GlobalChange', and the EU Horizon 2020 project 'HARMONIA'.} and Yifang Ban\textsuperscript{1}}

\address{
	\textsuperscript{1 }Division of Geoinformatics, KTH Royal Institute of Technology, 114 28 Stockholm, Sweden\\
}
\begin{document}
%
\maketitle
\begin{abstract}
This paper proposes a novel multi-temporal urban mapping approach using multi-modal satellite data from the Sentinel-1 Synthetic Aperture Radar (SAR) and Sentinel-2 MultiSpectral Instrument (MSI) missions. In particular, it focuses on the problem of a partly missing optical modality due to clouds. The proposed model utilizes two networks to extract features from each modality separately. In addition, a reconstruction network is utilized to approximate the optical features based on the SAR data in case of a missing optical modality. Our experiments on a multi-temporal urban mapping dataset with Sentinel-1 SAR and Sentinel-2 MSI data demonstrate that the proposed method outperforms a multi-modal approach that uses zero values as a replacement for missing optical data, as well as a uni-modal SAR-based approach. Therefore, the proposed method is effective in exploiting multi-modal data, if available, but it also retains its effectiveness in case the optical modality is missing.
\end{abstract}

\section{INTRODUCTION}

Multi-modal deep learning offers new opportunities for timely and accurate urban mapping and change detection by exploiting the complementary information acquired by Synthetic Aperture Radar (SAR) and optical sensors. In particular, the Copernicus Program's Sentinel-1 (S1) SAR and Sentinel-2 (S2) MultiSpectral Instrument (MSI) missions are playing a key role in multi-modal remote sensing research. For example, our previous work demonstrated that the complementary information in S1 SAR and S2 MSI data can be utilized to improve the transferability of deep learning models for urban extraction at a global scale \cite{hafner2022unsupervised}.

However, while the availability of multi-modal satellite data from S1 and S2 is a valid assumption for uni-temporal mapping, for multi-temporal urban mapping and change detection, the optical modality may not always be available due to cloud cover or other atmospheric conditions. Surprisingly, multi-modal research in the remote sensing domain has paid little attention to the missing modality problem, with few exceptions. For example, Saha \textit{et al.} \cite{saha2022supervised} considered the case of urban change detection with both optical and SAR images available at $t_1$, but only SAR images at $t_2$. The proposed Siamese network, accounting for the missing optical modality at $t_2$, outperformed a network using only SAR data. On the other hand, Zheng \textit{et al.} \cite{zheng2021deep} proposed a prototype network that learns a meta-sensory representation for all-weather building mapping. The prototype network is capable of dynamically generating sensor-specific networks that can be trained on multi-modal data. For SAR and optical data, the respective sensor-specific network generated by the prototype network outperformed networks that were directly trained on either modality. Recently, Li \textit{et al.} \cite{li2022dense} proposed an efficient dense adaptive grouping distillation network to effectively tackle multi-modal deep learning when one of the modalities is completely missing for inference.

Although these methods proved to be effective in tackling the missing modality problem, they are focusing on the case when a modality is completely missing for inference. In practice, however, a modality is often only partly missing. Therefore, it is desirable to develop methods that are capable of exploiting multi-modal data for inference, if both modalities are available. Bischke \textit{et al.} \cite{bischke2018overcoming} addressed this by generating missing depth images from optical images using Generative Adversarial Networks (GANs). However, there exist major challenges in the training of GANs, i.e., mode collapse, non-convergence, and instability. Moreover, SAR-to-optical image translation may be limited by the scene complexity and textural information \cite{zhao2022comparative}.

In this paper, we propose a multi-temporal urban mapping model with the capability to exploit the complementary information in SAR and optical data, while also being able to exploit uni-modal SAR data, in case the optical modality is missing. The model consists of two networks to extract features from the SAR and optical input separately, in addition to a third network that reconstructs the optical features from the SAR input in order to cope with a missing optical modality. The effectiveness of the proposed model is demonstrated on the multi-temporal urban mapping problem posed by the SpaceNet7 dataset \cite{van2021multi} using S1 SAR and S2 MSI data.

\section{Methodology}
\label{sec:methodology}

\subsection{Satellite Data Preparation}

\begin{figure}[h]
    \centering
    \includegraphics[width=.48\textwidth]{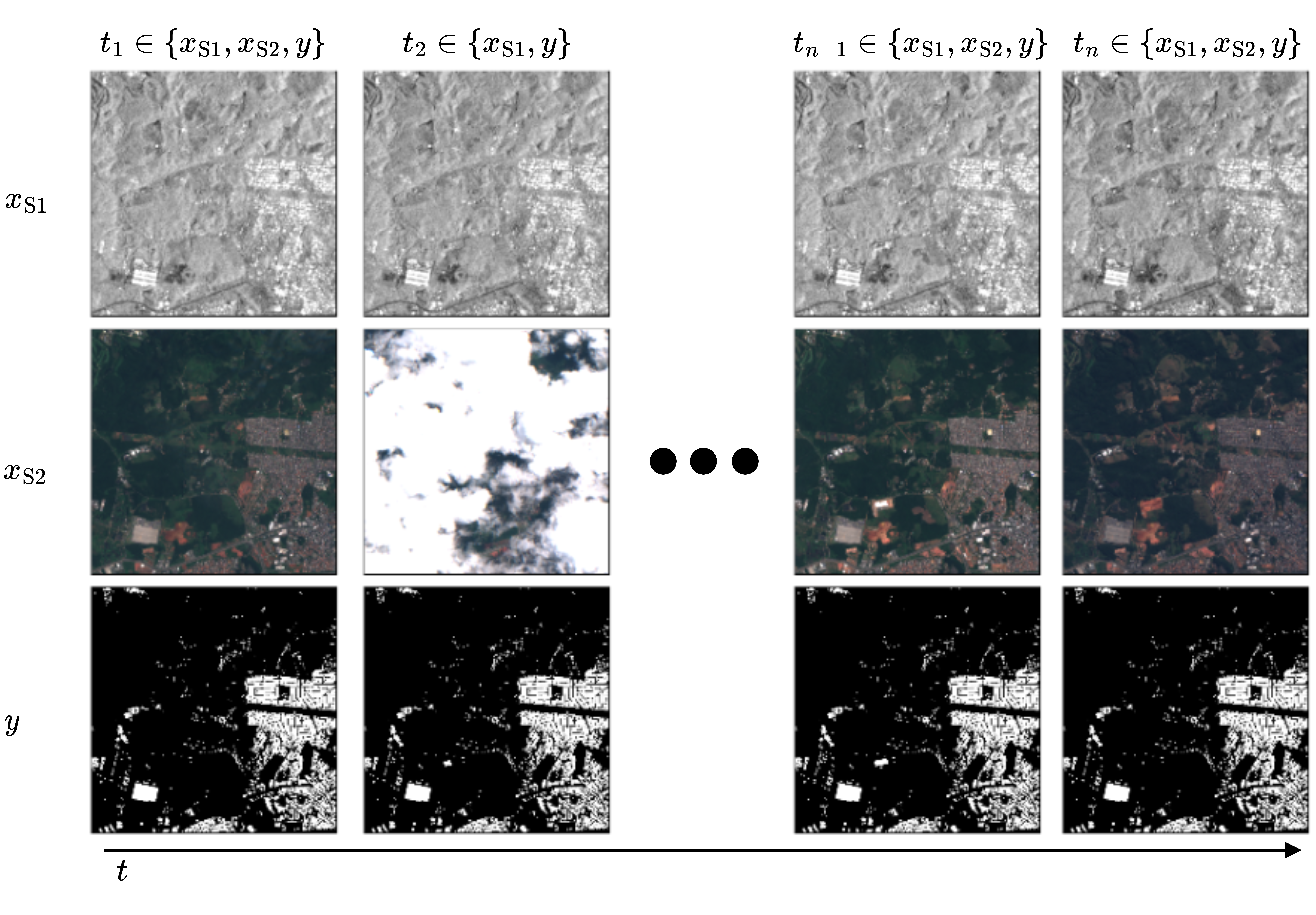}
    \caption{Data triples consisting of satellite data from S1 SAR and S2 MSI in addition to building labels ($\{x_{\rm S1}, x_{\rm S2}, y\}$) for a selection of timestamps $t$ in a time series of length $n$.}
    \label{fig:dataset_sample}
\end{figure}

The SpaceNet 7 dataset \cite{van2021multi} was leveraged to generate a multi-temporal urban mapping dataset featuring S1 SAR and S2 MSI images. It contains monthly time series of Planet imagery acquired between 2017 and 2020 and corresponding manually annotated building footprints for 60 distinct sites across the globe. S1 SAR images (VV + VH polarization at 20 m spatial resolution) and S2 MSI images (blue, green, red, and near-infrared at 10 m spatial resolution) were acquired for each of the approximately 24 timestamps per study site. The monthly S1 SAR image and S2 MSI image correspond to the temporal mean of all acquired S1 scenes within that month and the least cloudy S2 scene, respectively. For an in-depth description of the data preprocessing, we refer to \cite{hafner2022multi}, Finally, the images were resampled to the Ground Sampling Distance of the respective Planet images (GDS of approx. 4 m). Fig. \ref{fig:dataset_sample} exemplifies the prepared data triplets for a section of timestamps in a time series. It should be noted that the optical modality of the second timestamp ($t_2$) is considered missing due to clouds. The prepared data is available on Zenodo\footnote{\url{https://doi.org/10.5281/zenodo.7794693}}.

\subsection{Proposed Approach}

\begin{figure*}[ht]
    \centering
    \includegraphics[width=\textwidth]{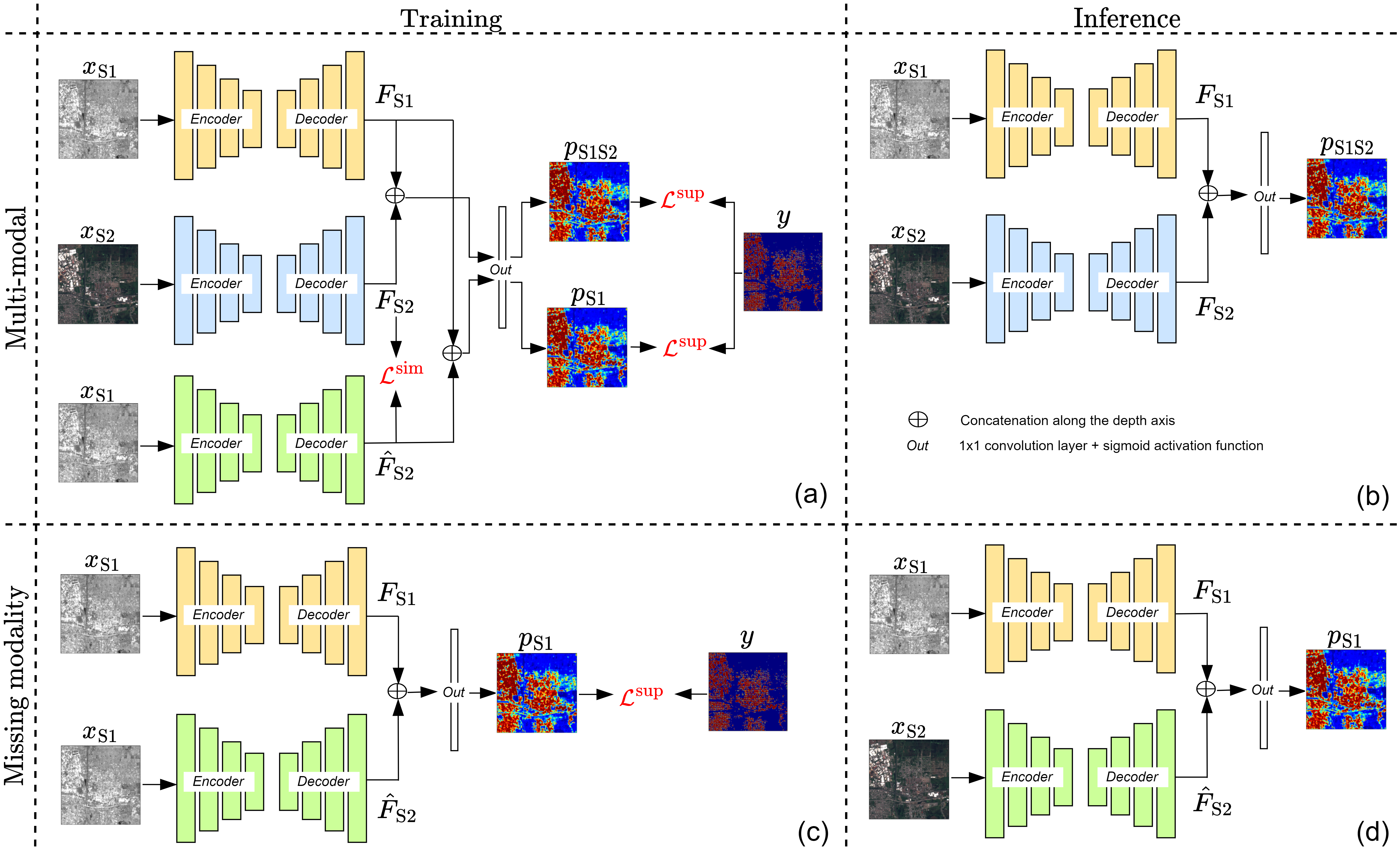}
    \caption{Overview of the proposed approach for multi-modal urban extraction with a partly missing optical modality.}
    \label{fig:overview_approach}
\end{figure*}

The proposed approach is illustrated in Fig. \ref{fig:overview_approach}. The model consists of two networks with identical architectures to separately extract multi-channel feature maps from the S1 SAR image ($F_{\rm S1}$) and the S2 MSI image ($F_{\rm S2}$). A prediction is then obtained from the concatenated feature maps in a late fusion manner. The popular encoder-decoder network U-Net \cite{ronneberger2015u} is used as underlying architecture for the two networks. To cope with missing S2 data, a feature reconstruction network is introduced. The reconstruction network uses the S1 image to approximate the multi-channel feature map extracted from the S2 image. The architecture of the reconstruction network is also based on U-Net.

The loss function differentiates between two cases: (1) the multi-modal case where both the S1 image and S2 image are available, and (2) the missing modality case where only the S1 image is available. In the multi-modal case, a prediction $p_{\rm S1S2}$ is obtained from the fused feature maps extracted from the S1 image and S2 image. The prediction is then used to compute a supervised loss term using label $y$. A second prediction $p_{\rm S1}$ is obtained from simulating a missing optical modality by replacing the feature map extracted from the S2 image ($F_{\rm S2}$) with its approximation ($\hat{F}_{\rm S2}$) produced by the reconstruction network from the S1 image. Accordingly, a second supervised loss term is computed based on $p_{\rm S1}$ and $y$. The third loss term is used to train the reconstruction network by minimizing the difference between the feature map obtained from the S2 image $F_{\rm S2}$ and its approximation $\hat{F}_{\rm S2}$ obtained from the reconstruction network. In the missing modality case, the loss function consists of a single supervised term. Having only the S1 image available, a prediction $p_{S1}$ is obtained from the features extracted by the S1 network and the approximated features obtained from the reconstruction network and compared to the corresponding label.

A Jaccard-like loss function, namely Power Jaccard loss, was used for supervised terms. In comparison to Jaccard loss, Power Jaccard accounts for continuous values by replacing the intersection and union in Intersection over Union (IoU) with product and sum, respectively, and it increases the weight of wrong predictions by introducing exponents in the denominator. On the other hand, L2 loss was used for the similarity term. It should be noted that the impact of the similarity term on the loss is regulated by hyper-parameter $\varphi$ in the multi-modal case.

During training, we use mini-batch gradient descent where a mini-batch can consist of multi-modal samples and uni-modal samples with a missing optical modality. Consequently, the cost for a mini-batch is computed by determining the loss for each sample in the mini-batch separately, before adding them together.

\subsection{Experimental Setup}

The 60 sites were split into a training, validation, and test set containing 41, 15, and 14 sites, respectively. The training set consists of 591 samples, each corresponding to a unique timestamp of the 31 training sites. For each epoch, training samples were generated dynamically by randomly cropping patches of size 64 x 64 pixels from the images and labels. Furthermore, flips and rotations were used as data augmentation to increase the diversity of the training set. AdamW was employed as an optimizer with an initial learning rate of $10^{-5}$. Hyper-parameter $\varphi$ was set to $10^{-2}$. Models were trained with a batch size of 16 for 100 epochs, but early stopping (patience 10) was employed to avoid overfitting. Everything was implemented using Facebook's deep learning framework PyTorch.

The proposed model was compared to two baselines: 1) U-Net \cite{ronneberger2015u} trained on S1 data and 2) a Dual Stream (DS) U-Net that fuses extracted features from a separate S1 and S2 network branch at the decision level, as proposed in \cite{hafner2021sentinel} for multi-modal urban change detection. Since the DS U-Net always requires two input modalities, an array of all zeros was used as a replacement for the optical modality in case only the SAR modality is available.

We used two common accuracy metrics for the quantitative accuracy assessment: F1 score and IoU. Formulas for the metrics are given in Eq.\ \ref{eq:accuracy_metrics}, where TP, FP and FN denote true positives, false positives, and false negatives, respectively.

\begin{equation}
\label{eq:accuracy_metrics}
    F1 = \frac{TP}{TP + \frac{1}{2}(FP + FN)}\; IoU = \frac{TP}{TP + FP + FN}
\end{equation}

\section{Results}
\label{sec:results}

Results were obtained for each model by training it five times with different seeds. Table \ref{tab:quantitative_results} lists F1 score and IoU ($\mu \pm \sigma$) for the three models. Column 1 summarizes the results for all of the test samples, Column 2 for only the test samples with S1 and S2 images available, and Column 3 for only the test samples with a missing modality. It should be noted that about 12 \% of the timestamps in the test set were affected by a missing optical modality. Higher mean values for both accuracy metrics were obtained by DS U-Net which uses multi-modal data if available compared to U-Net which exclusively uses S1 data. However, a closer look reveals that while the DS U-Net approach achieved improvements over the U-Net approach on the multi-modal samples (+ 0.062 F1 and + 0.049 IoU), its performance on the samples with a missing optical modality is considerably worse than that of the U-Net approach (- 0.115 F1 and - 0.078 IoU). In contrast to the DS U-Net approach, the proposed approach achieved better performance on the missing modality samples, although falling slightly short of the U-Net approach (- 0.021 F1 and - 0.015 IoU). Moreover, it clearly outperformed the DS U-Net and U-Net approaches on the multi-modal samples. Therefore, the highest overall accuracy values --- taking into account multi-modal and missing modality samples --- were obtained using the proposed approach.

\begin{table*}[ht]
  \caption{Quantitative test result. Values represent $\mu \pm \sigma$ of 5 runs. The highest mean values are boldfaced.}
  \label{tab:quantitative_results}
  \centering
  \begin{tabular}{lcccccccc}
    \toprule
    Method & \multicolumn{2}{c}{All samples} & \multicolumn{2}{c}{Multi-modal samples} & \multicolumn{2}{c}{Missing modality samples} \\
     & F1 $\uparrow$ & IoU $\uparrow$ & F1 $\uparrow$ & IoU $\uparrow$ & F1 $\uparrow$ & IoU $\uparrow$ \\
    \cmidrule(r){2-3} \cmidrule(r){4-5} \cmidrule(r){6-7}
    U-Net S1 & 0.362 $\pm$ 0.005 & 0.221 $\pm$ 0.003 & 0.364 $\pm$ 0.005 & 0.222 $\pm$ 0.004 & \textbf{0.348} $\pm$ \textbf{0.004} & \textbf{0.210} $\pm$ \textbf{0.003} \\
    DS U-Net & 0.411 $\pm$ 0.008 & 0.259 $\pm$ 0.006 & 0.426 $\pm$ 0.008 & 0.271 $\pm$ 0.006 & 0.233 $\pm$ 0.033 & 0.132 $\pm$ 0.021 \\
    Proposed & \textbf{0.423} $\pm$ \textbf{0.006} & \textbf{0.269} $\pm$ \textbf{0.005} & \textbf{0.435} $\pm$ \textbf{0.006} & \textbf{0.278} $\pm$ \textbf{0.005} & 0.327 $\pm$ 0.008 & 0.195 $\pm$ 0.006 \\
    \bottomrule
  \end{tabular}
\end{table*}

Finally, it should be noted that the obtained accuracy values in Table \ref{tab:quantitative_results} are relatively low, especially compared to state-of-the-art urban mapping approaches using multi-modal S1 SAR and S2 MSI data (e.g. \cite{hafner2022unsupervised}). However, unlike in most land cover mapping studies including \cite{hafner2022unsupervised}, model predictions were assessed at a finer spatial resolution (i.e., approx. 4 m) than that of the input imagery (i.e., 10 m and 20 m for S1 and S2 imagery, respectively). This decision was made to maintain correspondence with the Planet imagery. Another reason for the low accuracy values is that the SpaceNet-7 sites are very challenging since many of them are located in arid and semi-arid areas that are undergoing rapid urbanization.

\FloatBarrier

\section{Discussion}
\label{sec:discussion}

The quantitative results (Table \ref{tab:quantitative_results}) highlight that a multi-modal approach dealing with a partly missing modality by replacing it with zero values outperforms a uni-modal approach on an urban mapping dataset consisting of predominately multi-modal samples (approx. 90 \%). However, the results also reveal that this way of dealing with missing modality samples leads to bad performance on uni-modal samples. The proposed approach successfully mitigates this effect by employing a separate network to reconstruct S2 features from S1 input images. While the proposed approach was deemed effective to deal with a partly missing modality in this study, it should also be emphasized that its performance on uni-modal samples is lower than that of a uni-modal network.

\section{Conclusion}
\label{sec:conclusion}

This paper addressed the problem of a partly missing optical modality during inference time for multi-modal urban mapping from S1 and S2 data. To that end, a model that utilizes a reconstruction network to approximate the features of the optical modality when its missing was proposed. The model achieved improved performance over a uni-modal approach trained on S1 SAR
data and a multi-modal approach using zero values in case of a missing optical modality. Despite these improvements, our findings indicate that further research is needed to improve the performance of missing modality samples.

\FloatBarrier
\bibliography{ref.bib}
\bibliographystyle{IEEEbib}

\end{document}